%% file: main.tex
\documentclass[10pt,twocolumn,letterpaper]{article}

\usepackage{cvpr}              
\input{preamble}
\definecolor{cvprblue}{rgb}{0.21,0.49,0.74}
\usepackage[pagebackref,breaklinks,colorlinks,allcolors=cvprblue]{hyperref}

\def\paperID{XAI4CV-25} 
\def\confName{CVPR}
\def\confYear{2026}

\title{Feature Attribution Stability Suite: How Stable Are Post-Hoc Attributions?
}

\author{Kamalasankari Subramaniakuppusamy\\
The George Washington University\\
Washington D.C.\\
{\tt\small kamalasankaris@gwu.edu}
\and
Jugal Gajjar\\
The George Washington University\\
Washington D.C.\\
{\tt\small jugal.gajjar@gwu.edu}
}

\begin{document}
\maketitle
\input{sec/0_abstract}    
\input{sec/1_intro}
\input{sec/2_litrev}
\input{sec/3_methodology}
\input{sec/4_exp-and-results}
\input{sec/5_conclusion}
\newpage
{
    \small
    \bibliographystyle{ieeenat_fullname}
    \bibliography{main}
}

\end{document}


\twocolumn[
\begin{center}
    {\Large\bfseries Feature Attribution Stability Suite: How Stable Are Post-Hoc Attributions?\par}
    \vspace{0.4em}
    {\large Supplementary Material\par}
    \vspace{1.5em}
\end{center}
]

\appendix
\noindent{\large\bfseries Appendix}
\vspace{0.4em}

\noindent In the Appendix, we provide the following:
\begin{itemize}[leftmargin=1.5em, itemsep=2pt]
\item metric design details, including Spearman rescaling
      and computational cost breakdown
      (Appendix~\ref{sec:metric-details}).
\item per-perturbation stability scores, retention rates,
      and heatmaps for CIFAR-10
      (Appendix~\ref{sec:cifar10}), ImageNet
      (Appendix~\ref{sec:imagenet}), and COCO
      (Appendix~\ref{sec:coco}).
\end{itemize}
\noindent All values are computed over prediction-invariant image pairs. Scores in conditions with near-zero retention ($<$0.1\%) rest on very few evaluation pairs and should be interpreted with caution; such cells are marked with $^\dagger$ in the tables below.

\section{Metric Design Details}\label{sec:metric-details}
\subsection{Spearman Rescaling}
Raw Spearman rank correlation~\cite{spearman} $\rho_s$ ranges over
$[-1, +1]$, where $+1$ indicates identical importance ordering, $0$
indicates no linear rank association, and $-1$ indicates perfectly
reversed ordering. SSIM~\cite{wang} and top-$k$ Jaccard~\cite{jaccard}
both range over $[0, 1]$. To ensure equal contribution to the composite
FASS score, we rescale Spearman via $R = (\rho_s + 1)/2$, mapping
perfect agreement to $1.0$, random ordering to $0.5$, and complete
reversal to $0.0$. Without this rescaling, negative Spearman values
would pull the unweighted mean disproportionately downward relative to
the other two components, which cannot take negative values.

\subsection{Computational Cost}

LIME was the most expensive method, accounting for approximately 70\%
of total runtime ($\sim$ 100--120 GPU hours) due to its
sampling-based surrogate modeling. GradientSHAP required approximately
48 GPU hours owing to background sampling and gradient
accumulation. Gradient-based methods (Integrated Gradients and
Grad-CAM) were substantially cheaper, collectively requiring fewer than 30 GPU hours. Peak GPU memory usage occurred during
GradientSHAP evaluations ($\sim$38.2\,GB), approaching the 40\,GB
hardware limit. Given Colab session limits and runtime interruptions,
effective wall-clock duration extended to several months.

\section{CIFAR-10}\label{sec:cifar10}

Table~\ref{tab:cifar10-ret} reports prediction-invariant retention rates for all architecture--perturbation combinations on CIFAR-10. These rates are identical across attribution methods (retention depends only on the model and perturbation, not the explanation technique). Row-wise averages of the stability tables below correspond to the per-model scores in Table~4 of the main paper.

\begin{table}[H]
\centering
\scriptsize
\caption{Prediction-invariant retention (\%) on CIFAR-10.}
\label{tab:cifar10-ret}
\begin{tabular}{@{}l ccccc@{}}
\toprule
\textbf{Architecture} & \textbf{Rot.} & \textbf{Trans.} & \textbf{Bright.} & \textbf{Noise} & \textbf{JPEG} \\
\midrule
ResNet-50    & 37.2 & 0.6 & 9.0 & 26.9 & $<$0.1 \\
DenseNet-121 & 20.8 & $<$0.1 & $<$0.1 & 12.1 & $<$0.1 \\
ConvNeXt-T   & 45.3 & $<$0.1 & $<$0.1 & 11.6 & $<$0.1 \\
ViT-B/16     & 36.1 & $<$0.1 & $<$0.1 & 29.2 & $<$0.1 \\
\bottomrule
\end{tabular}
\end{table}

\subsection{Integrated Gradients}

\begin{table}[H]
\centering
\scriptsize
\setlength{\tabcolsep}{2.5pt}
\begin{tabular}{@{}l ccccc@{}}
\toprule
\textbf{Architecture} & \textbf{Rot.} & \textbf{Trans.}$^\dagger$ & \textbf{Bright.}$^\dagger$ & \textbf{Noise} & \textbf{JPEG}$^\dagger$ \\
\midrule
ResNet-50    & 0.422 & 0.560 & 0.470 & 0.381 & 0.528 \\
DenseNet-121 & 0.446 & 0.532 & 0.511 & 0.380 & 0.599 \\
ConvNeXt-T   & 0.456 & 0.581 & 0.607 & 0.439 & 0.652 \\
ViT-B/16     & 0.440 & 0.542 & 0.569 & 0.470 & 0.650 \\
\bottomrule
\end{tabular}
\caption{IG stability on CIFAR-10. $^\dagger$Near-zero retention; interpret with caution.}
\label{tab:cifar10-ig}
\end{table}

\begin{figure}[H]
\centering
\includegraphics[width=0.95\linewidth]{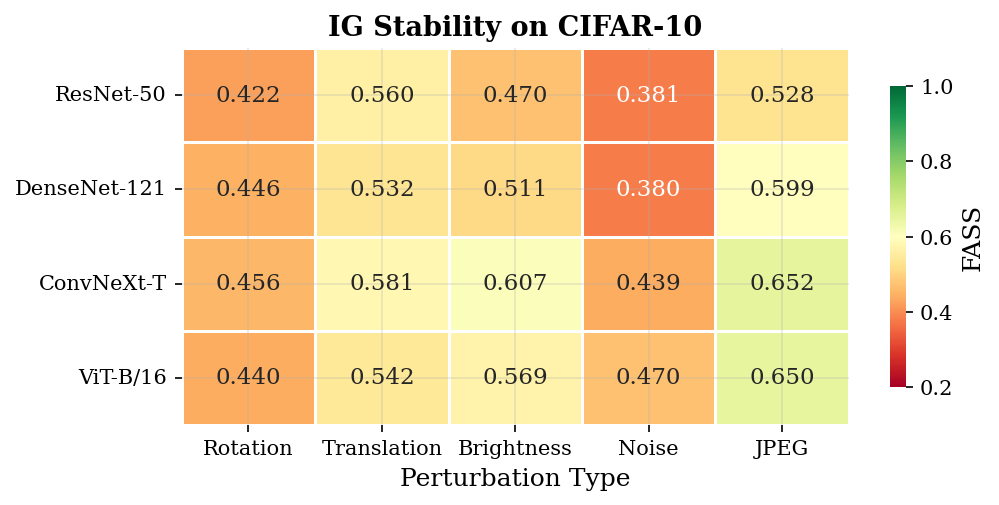}
\caption{IG stability on CIFAR-10.}
\label{fig:cifar10-ig}
\end{figure}

\subsection{GradientSHAP}

\begin{table}[H]
\centering
\scriptsize
\setlength{\tabcolsep}{2.5pt}
\begin{tabular}{@{}l ccccc@{}}
\toprule
\textbf{Architecture} & \textbf{Rot.} & \textbf{Trans.}$^\dagger$ & \textbf{Bright.}$^\dagger$ & \textbf{Noise} & \textbf{JPEG}$^\dagger$ \\
\midrule
ResNet-50    & 0.423 & 0.501 & 0.425 & 0.393 & 0.468 \\
DenseNet-121 & 0.445 & 0.491 & 0.479 & 0.384 & 0.520 \\
ConvNeXt-T   & 0.457 & 0.536 & 0.563 & 0.441 & 0.583 \\
ViT-B/16     & 0.442 & 0.517 & 0.523 & 0.463 & 0.588 \\
\bottomrule
\end{tabular}
\caption{GradientSHAP stability on CIFAR-10. $^\dagger$Near-zero retention.}
\label{tab:cifar10-shap}
\end{table}

\begin{figure}[H]
\centering
\includegraphics[width=0.95\linewidth]{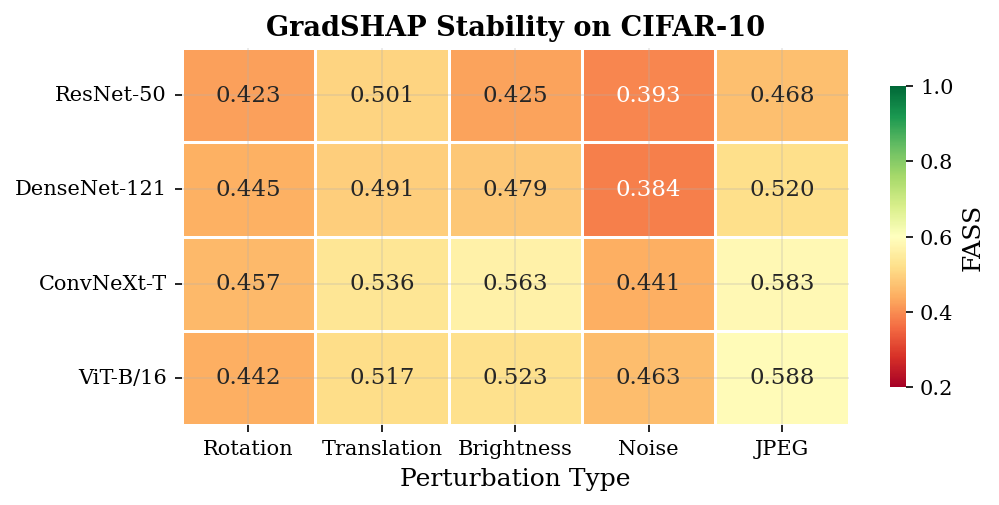}
\caption{GradientSHAP stability on CIFAR-10.}
\label{fig:cifar10-shap}
\end{figure}

\subsection{Grad-CAM}

\begin{table}[H]
\centering
\scriptsize
\setlength{\tabcolsep}{2.5pt}
\begin{tabular}{@{}l ccccc@{}}
\toprule
\textbf{Architecture} & \textbf{Rot.} & \textbf{Trans.}$^\dagger$ & \textbf{Bright.}$^\dagger$ & \textbf{Noise} & \textbf{JPEG}$^\dagger$ \\
\midrule
ResNet-50    & 0.450 & 0.529 & 0.643 & 0.449 & 0.703 \\
DenseNet-121 & 0.535 & 0.523 & 0.735 & 0.553 & 0.760 \\
ConvNeXt-T   & 0.548 & 0.834 & 0.796 & 0.645 & 0.774 \\
ViT-B/16     & 0.539 & 0.667 & 0.744 & 0.598 & 0.722 \\
\bottomrule
\end{tabular}
\caption{Grad-CAM stability on CIFAR-10. $^\dagger$Near-zero retention.}
\label{tab:cifar10-gradcam}
\end{table}

\begin{figure}[H]
\centering
\includegraphics[width=0.95\linewidth]{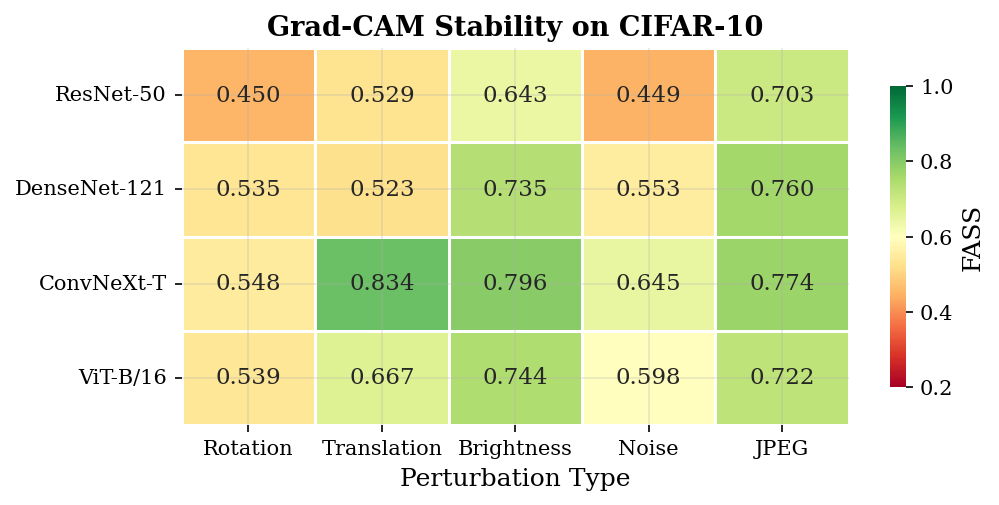}
\caption{Grad-CAM stability on CIFAR-10.}
\label{fig:cifar10-gradcam}
\end{figure}

\subsection{LIME}

\begin{table}[H]
\centering
\scriptsize
\setlength{\tabcolsep}{2.5pt}
\begin{tabular}{@{}l ccccc@{}}
\toprule
\textbf{Architecture} & \textbf{Rot.} & \textbf{Trans.}$^\dagger$ & \textbf{Bright.}$^\dagger$ & \textbf{Noise} & \textbf{JPEG}$^\dagger$ \\
\midrule
ResNet-50    & 0.282 & 0.346 & 0.346 & 0.284 & 0.350 \\
DenseNet-121 & 0.279 & 0.339 & 0.338 & 0.273 & 0.344 \\
ConvNeXt-T   & 0.295 & 0.362 & 0.356 & 0.288 & 0.364 \\
ViT-B/16     & 0.280 & 0.368 & 0.357 & 0.287 & 0.437 \\
\bottomrule
\end{tabular}
\caption{LIME stability on CIFAR-10. $^\dagger$Near-zero retention.}
\label{tab:cifar10-lime}
\end{table}

\begin{figure}[H]
\centering
\includegraphics[width=0.95\linewidth]{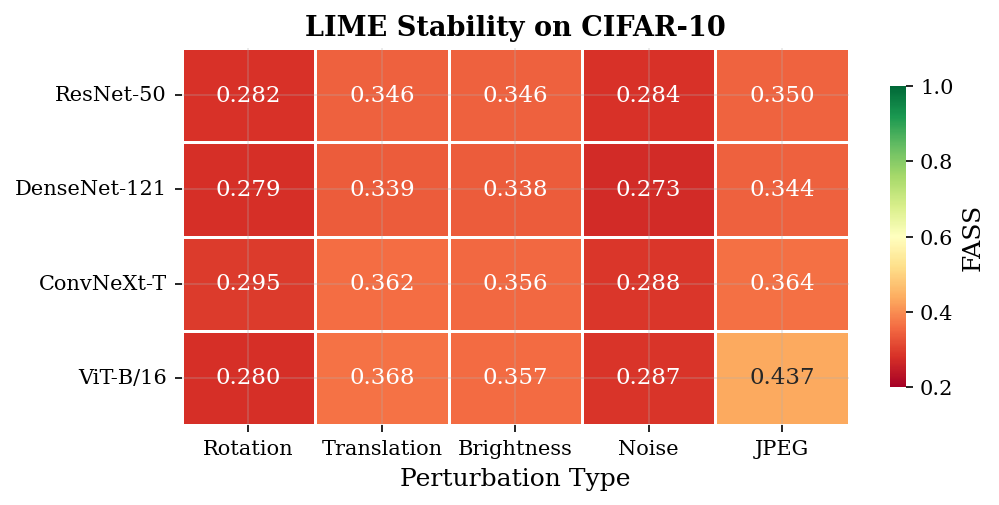}
\caption{LIME stability on CIFAR-10.}
\label{fig:cifar10-lime}
\end{figure}

\paragraph{Summary.}
CIFAR-10 yields the lowest stability across all datasets, consistent with distribution mismatch from upsampling $32 \times 32$ inputs to $224 \times 224$. Grad-CAM achieves the highest stability across all perturbations and architectures. IG and GradientSHAP track closely, reflecting their shared gradient-based formulation. LIME exhibits the lowest stability overall. ConvNeXt-T and ViT-B/16 show slightly improved stability compared to ResNet-50 and DenseNet-121, though differences remain moderate.

\section{ImageNet}\label{sec:imagenet}

Table~\ref{tab:imagenet-ret} reports retention rates on ImageNet. Only rotation and noise retain substantial prediction-invariant pairs; translation, brightness, and JPEG yield near-zero retention across all architectures. Row-wise averages of the stability tables below correspond to the per-model scores in Table~4 of the main paper.

\begin{table}[H]
\centering
\scriptsize
\caption{Prediction-invariant retention (\%) on ImageNet.}
\label{tab:imagenet-ret}
\begin{tabular}{@{}l ccccc@{}}
\toprule
\textbf{Architecture} & \textbf{Rot.} & \textbf{Trans.} & \textbf{Bright.} & \textbf{Noise} & \textbf{JPEG} \\
\midrule
ResNet-50    & 58.8 & $<$0.1 & $<$0.1 & 70.1 & $<$0.1 \\
DenseNet-121 & 58.4 & $<$0.1 & $<$0.1 & 70.5 & $<$0.1 \\
ConvNeXt-T   & 68.3 & $<$0.1 & $<$0.1 & 71.3 & $<$0.1 \\
ViT-B/16     & 35.7 & $<$0.1 & $<$0.1 & 73.0 & $<$0.1 \\
\bottomrule
\end{tabular}
\end{table}

\subsection{Integrated Gradients}

\begin{table}[H]
\centering
\scriptsize
\setlength{\tabcolsep}{2.5pt}
\begin{tabular}{@{}l ccccc@{}}
\toprule
\textbf{Architecture} & \textbf{Rot.} & \textbf{Trans.}$^\dagger$ & \textbf{Bright.}$^\dagger$ & \textbf{Noise} & \textbf{JPEG}$^\dagger$ \\
\midrule
ResNet-50    & 0.471 & 0.418 & 0.502 & 0.642 & 0.541 \\
DenseNet-121 & 0.466 & 0.409 & 0.498 & 0.642 & 0.538 \\
ConvNeXt-T   & 0.483 & 0.432 & 0.521 & 0.613 & 0.562 \\
ViT-B/16     & 0.497 & 0.448 & 0.553 & 0.768 & 0.589 \\
\bottomrule
\end{tabular}
\caption{IG stability on ImageNet. $^\dagger$Near-zero retention.}
\label{tab:imagenet-ig}
\end{table}

\begin{figure}[H]
\centering
\includegraphics[width=0.95\linewidth]{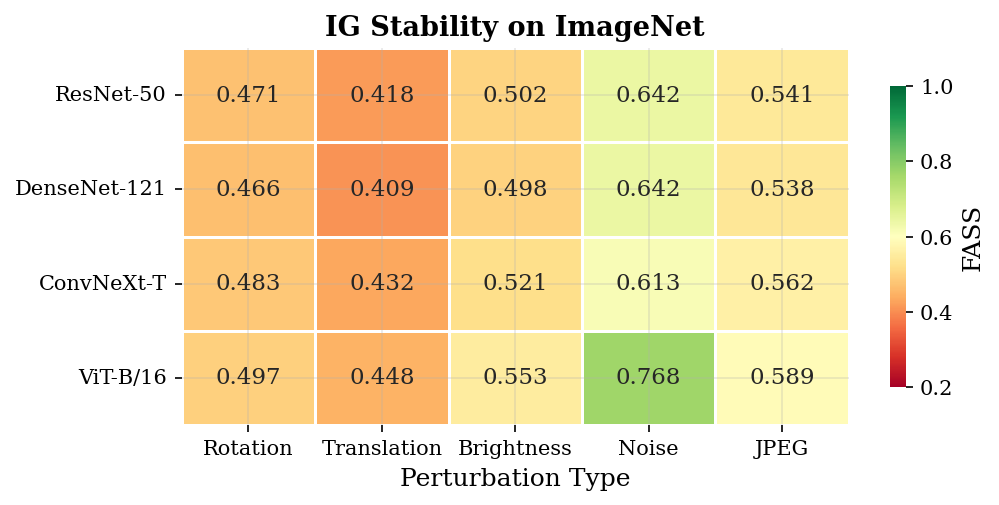}
\caption{IG stability on ImageNet.}
\label{fig:imagenet-ig}
\end{figure}

\subsection{GradientSHAP}

\begin{table}[H]
\centering
\scriptsize
\setlength{\tabcolsep}{2.5pt}
\begin{tabular}{@{}l ccccc@{}}
\toprule
\textbf{Architecture} & \textbf{Rot.} & \textbf{Trans.}$^\dagger$ & \textbf{Bright.}$^\dagger$ & \textbf{Noise} & \textbf{JPEG}$^\dagger$ \\
\midrule
ResNet-50    & 0.470 & 0.421 & 0.472 & 0.572 & 0.492 \\
DenseNet-121 & 0.465 & 0.412 & 0.468 & 0.572 & 0.487 \\
ConvNeXt-T   & 0.484 & 0.438 & 0.501 & 0.568 & 0.519 \\
ViT-B/16     & 0.490 & 0.443 & 0.512 & 0.626 & 0.531 \\
\bottomrule
\end{tabular}
\caption{GradientSHAP stability on ImageNet. $^\dagger$Near-zero retention.}
\label{tab:imagenet-shap}
\end{table}

\begin{figure}[H]
\centering
\includegraphics[width=0.95\linewidth]{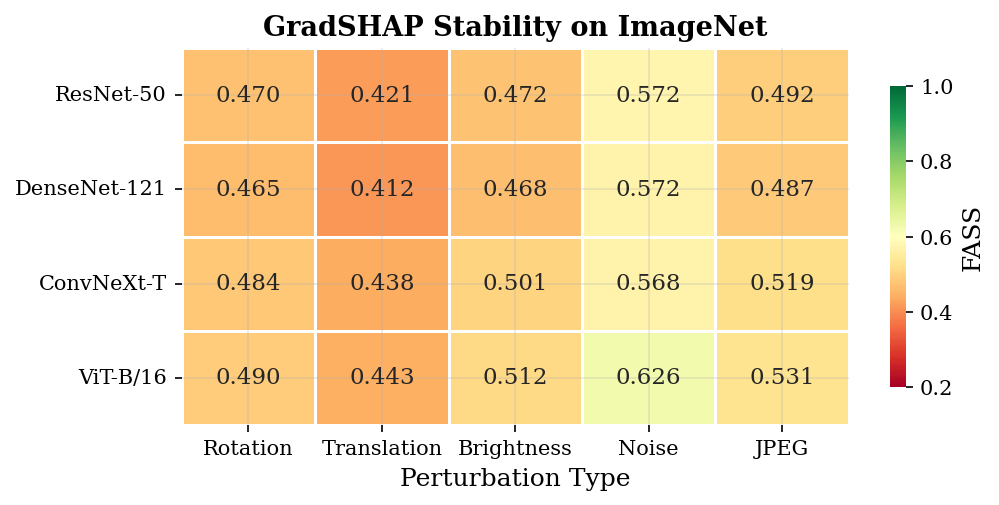}
\caption{GradientSHAP stability on ImageNet.}
\label{fig:imagenet-shap}
\end{figure}

\subsection{Grad-CAM}

\begin{table}[H]
\centering
\scriptsize
\setlength{\tabcolsep}{2.5pt}
\begin{tabular}{@{}l ccccc@{}}
\toprule
\textbf{Architecture} & \textbf{Rot.} & \textbf{Trans.}$^\dagger$ & \textbf{Bright.}$^\dagger$ & \textbf{Noise} & \textbf{JPEG}$^\dagger$ \\
\midrule
ResNet-50    & 0.726 & 0.671 & 0.821 & 0.877 & 0.853 \\
DenseNet-121 & 0.761 & 0.703 & 0.873 & 0.915 & 0.889 \\
ConvNeXt-T   & 0.779 & 0.724 & 0.869 & 0.897 & 0.878 \\
ViT-B/16     & 0.762 & 0.712 & 0.842 & 0.822 & 0.852 \\
\bottomrule
\end{tabular}
\caption{Grad-CAM stability on ImageNet. $^\dagger$Near-zero retention.}
\label{tab:imagenet-gradcam}
\end{table}

\begin{figure}[H]
\centering
\includegraphics[width=0.95\linewidth]{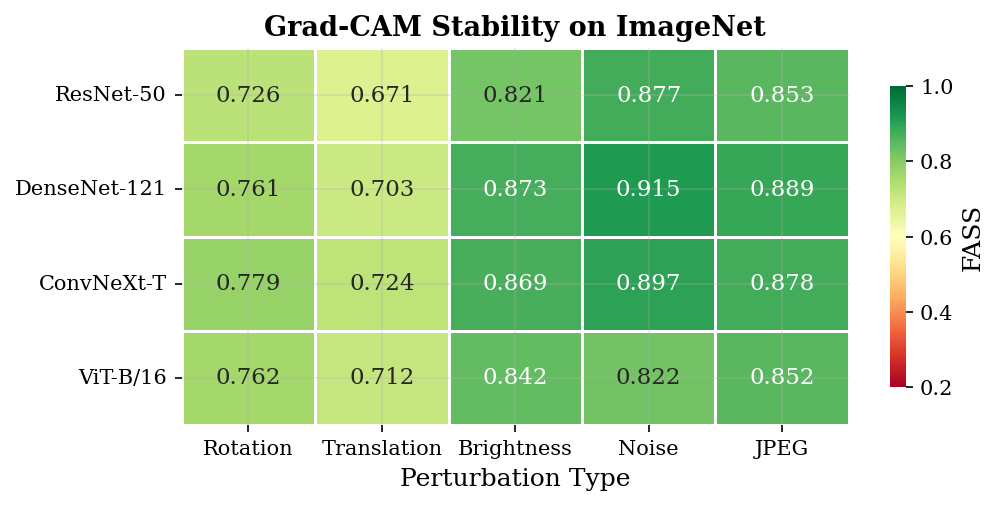}
\caption{Grad-CAM stability on ImageNet.}
\label{fig:imagenet-gradcam}
\end{figure}

\subsection{LIME}

\begin{table}[H]
\centering
\scriptsize
\setlength{\tabcolsep}{2.5pt}
\begin{tabular}{@{}l ccccc@{}}
\toprule
\textbf{Architecture} & \textbf{Rot.} & \textbf{Trans.}$^\dagger$ & \textbf{Bright.}$^\dagger$ & \textbf{Noise} & \textbf{JPEG}$^\dagger$ \\
\midrule
ResNet-50    & 0.410 & 0.340 & 0.390 & 0.470 & 0.430 \\
DenseNet-121 & 0.370 & 0.360 & 0.410 & 0.490 & 0.450 \\
ConvNeXt-T   & 0.390 & 0.350 & 0.430 & 0.470 & 0.440 \\
ViT-B/16     & 0.400 & 0.380 & 0.440 & 0.570 & 0.460 \\
\bottomrule
\end{tabular}
\caption{LIME stability on ImageNet. $^\dagger$Near-zero retention.}
\label{tab:imagenet-lime}
\end{table}

\begin{figure}[H]
\centering
\includegraphics[width=0.95\linewidth]{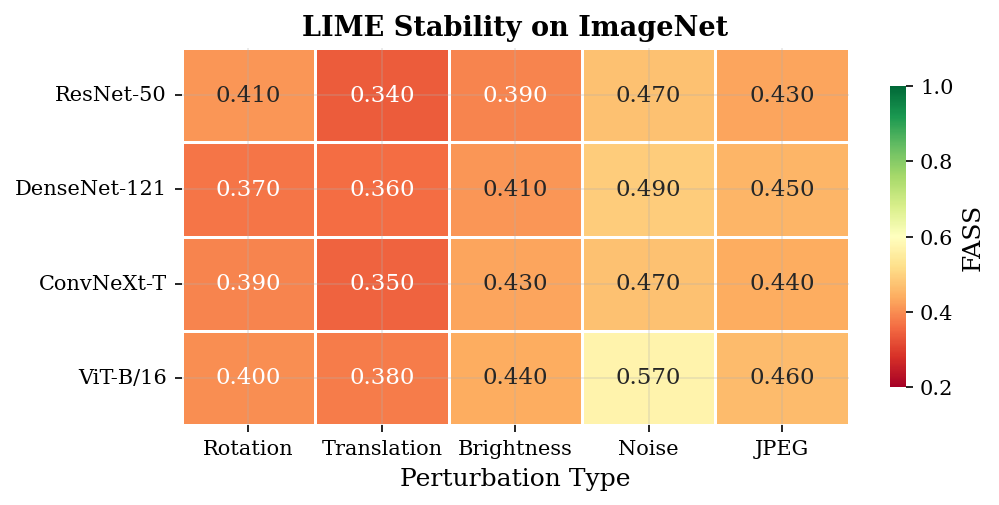}
\caption{LIME stability on ImageNet.}
\label{fig:imagenet-lime}
\end{figure}

\paragraph{Summary.}
ImageNet yields the highest stability among all three datasets for conditions with substantial retention (rotation and noise), consistent with pretrained feature alignment and native input resolution. Grad-CAM exceeds 0.85 under additive noise across all architectures. IG and GradientSHAP remain closely aligned. Translation, brightness, and JPEG produce near-zero retention, limiting the reliability of stability scores for those conditions.

\section{COCO}\label{sec:coco}

Table 2 in the main paper reports prediction-invariant retention rates by perturbation type. Rotation and noise retain substantial invariant pairs; JPEG retains a small fraction on ResNet-50 (11.7\%). Row-wise averages of the stability tables below correspond to the per-model scores in Table~4 of the main paper.

\begin{table}[H]
\centering
\scriptsize
\caption{Prediction-invariant retention (\%) on COCO.}
\label{tab:coco-ret}
\begin{tabular}{@{}l ccccc@{}}
\toprule
\textbf{Architecture} & \textbf{Rot.} & \textbf{Trans.} & \textbf{Bright.} & \textbf{Noise} & \textbf{JPEG} \\
\midrule
ResNet-50    & 88.1 & $<$0.1 & $<$0.1 & 74.3 & 11.7 \\
DenseNet-121 & 51.7 & $<$0.1 & $<$0.1 & 90.6 & $<$0.1 \\
ConvNeXt-T   & 46.1 & $<$0.1 & $<$0.1 & 62.8 & $<$0.1 \\
ViT-B/16     & 45.4 & $<$0.1 & $<$0.1 & 76.7 & 0.3 \\
\bottomrule
\end{tabular}
\end{table}

\subsection{Integrated Gradients}

\begin{table}[H]
\centering
\scriptsize
\setlength{\tabcolsep}{2.5pt}
\begin{tabular}{@{}l ccccc@{}}
\toprule
\textbf{Architecture} & \textbf{Rot.} & \textbf{Trans.}$^\dagger$ & \textbf{Bright.}$^\dagger$ & \textbf{Noise} & \textbf{JPEG} \\
\midrule
ResNet-50    & 0.446 & 0.358 & 0.472 & 0.606 & 0.405 \\
DenseNet-121 & 0.440 & 0.340 & 0.516 & 0.585 & 0.377$^\dagger$ \\
ConvNeXt-T   & 0.474 & 0.390 & 0.548 & 0.597 & 0.416$^\dagger$ \\
ViT-B/16     & 0.481 & 0.405 & 0.558 & 0.755 & 0.529 \\
\bottomrule
\end{tabular}
\caption{IG stability on COCO. $^\dagger$Near-zero retention.}
\label{tab:coco-ig}
\end{table}

\begin{figure}[H]
\centering
\includegraphics[width=0.95\linewidth]{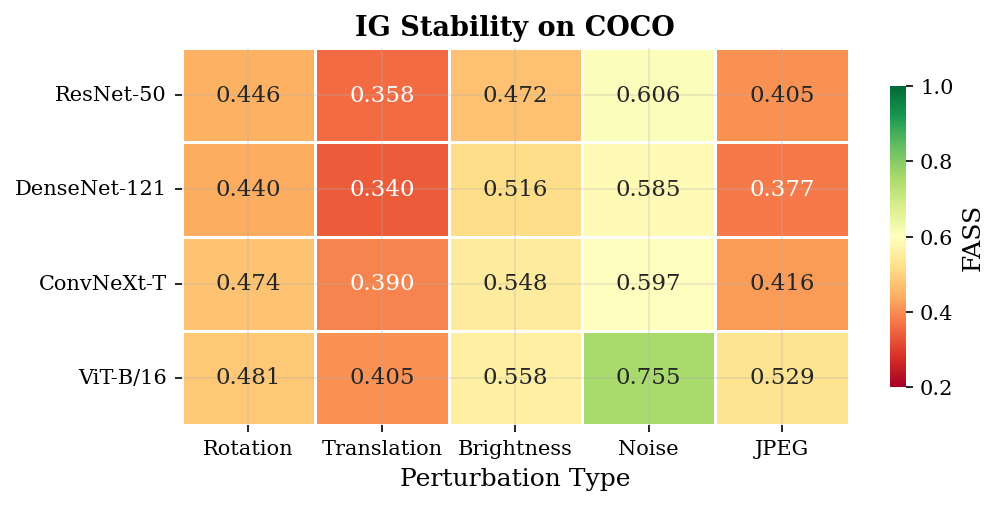}
\caption{IG stability on COCO.}
\label{fig:coco-ig}
\end{figure}

\subsection{GradientSHAP}

\begin{table}[H]
\centering
\scriptsize
\setlength{\tabcolsep}{2.5pt}
\begin{tabular}{@{}l ccccc@{}}
\toprule
\textbf{Architecture} & \textbf{Rot.} & \textbf{Trans.}$^\dagger$ & \textbf{Bright.}$^\dagger$ & \textbf{Noise} & \textbf{JPEG} \\
\midrule
ResNet-50    & 0.445 & 0.364 & 0.425 & 0.533 & 0.391 \\
DenseNet-121 & 0.437 & 0.339 & 0.442 & 0.519 & 0.367$^\dagger$ \\
ConvNeXt-T   & 0.475 & 0.393 & 0.506 & 0.568 & 0.411$^\dagger$ \\
ViT-B/16     & 0.481 & 0.404 & 0.515 & 0.647 & 0.494 \\
\bottomrule
\end{tabular}
\caption{GradientSHAP stability on COCO. $^\dagger$Near-zero retention.}
\label{tab:coco-shap}
\end{table}

\begin{figure}[H]
\centering
\includegraphics[width=0.95\linewidth]{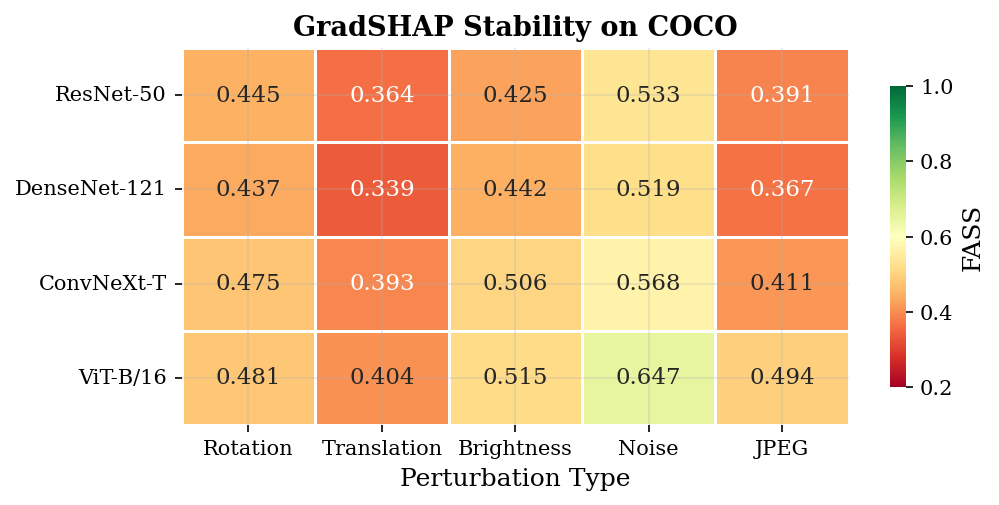}
\caption{GradientSHAP stability on COCO.}
\label{fig:coco-shap}
\end{figure}

\subsection{Grad-CAM}

\begin{table}[H]
\centering
\scriptsize
\setlength{\tabcolsep}{2.5pt}
\begin{tabular}{@{}l ccccc@{}}
\toprule
\textbf{Architecture} & \textbf{Rot.} & \textbf{Trans.}$^\dagger$ & \textbf{Bright.}$^\dagger$ & \textbf{Noise} & \textbf{JPEG} \\
\midrule
ResNet-50    & 0.645 & 0.609 & 0.676 & 0.825 & 0.662 \\
DenseNet-121 & 0.661 & 0.595 & 0.824 & 0.829 & 0.707$^\dagger$ \\
ConvNeXt-T   & 0.692 & 0.545 & 0.806 & 0.838 & 0.582$^\dagger$ \\
ViT-B/16     & 0.752 & 0.716 & 0.637 & 0.849 & 0.658 \\
\bottomrule
\end{tabular}
\caption{Grad-CAM stability on COCO. $^\dagger$Near-zero retention.}
\label{tab:coco-gradcam}
\end{table}

\begin{figure}[H]
\centering
\includegraphics[width=0.95\linewidth]{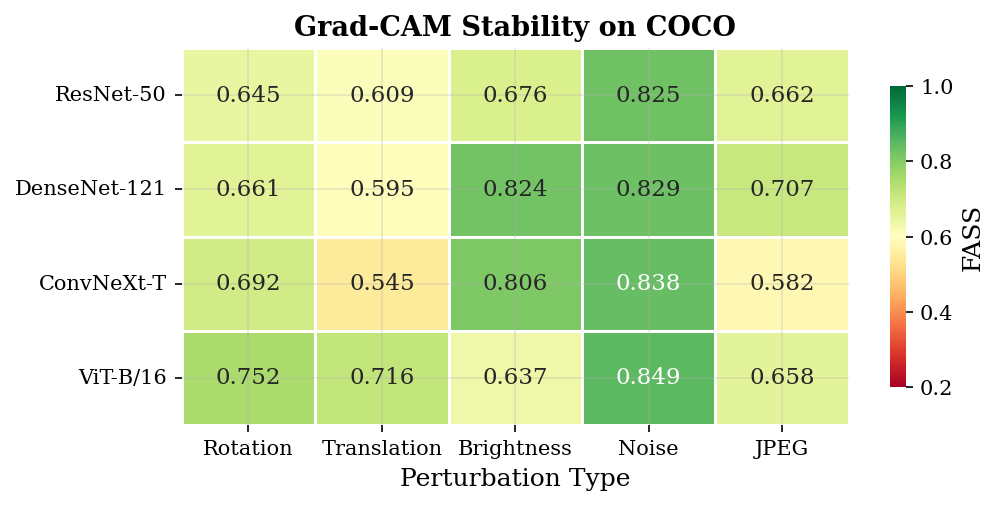}
\caption{Grad-CAM stability on COCO.}
\label{fig:coco-gradcam}
\end{figure}

\subsection{LIME}

\begin{table}[H]
\centering
\scriptsize
\setlength{\tabcolsep}{2.5pt}
\begin{tabular}{@{}l ccccc@{}}
\toprule
\textbf{Architecture} & \textbf{Rot.} & \textbf{Trans.}$^\dagger$ & \textbf{Bright.}$^\dagger$ & \textbf{Noise} & \textbf{JPEG} \\
\midrule
ResNet-50    & 0.427 & 0.489 & 0.529 & 0.415 & 0.512 \\
DenseNet-121 & 0.427 & 0.466 & 0.526 & 0.432 & 0.495$^\dagger$ \\
ConvNeXt-T   & 0.448 & 0.492 & 0.556 & 0.422 & 0.476$^\dagger$ \\
ViT-B/16     & 0.429 & 0.445 & 0.517 & 0.413 & 0.489 \\
\bottomrule
\end{tabular}
\caption{LIME stability on COCO. $^\dagger$Near-zero retention.}
\label{tab:coco-lime}
\end{table}

\begin{figure}[H]
\centering
\includegraphics[width=0.95\linewidth]{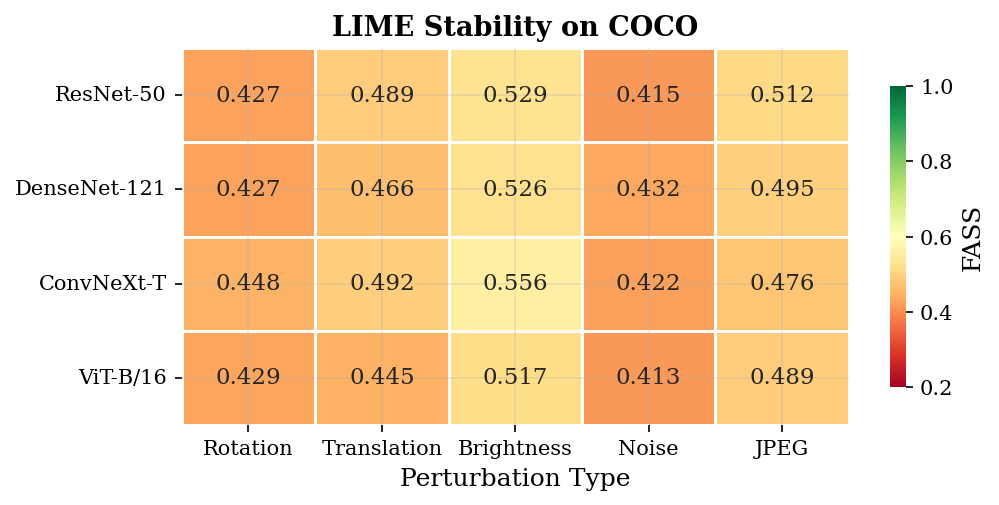}
\caption{LIME stability on COCO.}
\label{fig:coco-lime}
\end{figure}

\paragraph{Summary.}
COCO displays intermediate stability between CIFAR-10 and ImageNet. Grad-CAM remains the most stable method, though the gap with IG narrows relative to ImageNet. LIME exhibits stronger performance on COCO than on CIFAR-10, consistent with the hypothesis that multi-object scenes constrain perturbation-based sampling variability. Architectural differences are more pronounced under noise, where ViT-B/16 demonstrates elevated stability relative to convolutional architectures.

\newpage
\renewcommand{\refname}{Appendix References}

%% file: sec/0_abstract.tex
\begin{abstract}
\label{sec:abstract}
Post-hoc feature attribution methods are widely deployed in
safety-critical vision systems, yet their stability under realistic input perturbations remains poorly characterized. Existing metrics evaluate explanations primarily under additive noise, collapse stability to a single scalar, and fail to condition on prediction preservation, conflating explanation fragility with model sensitivity. We introduce the Feature Attribution Stability Suite (FASS), a benchmark that enforces prediction-invariance filtering, decomposes stability into three complementary metrics: structural similarity, rank correlation, and top-$k$ Jaccard overlap–and evaluates across geometric, photometric, and compression perturbations. Evaluating four attribution methods (Integrated Gradients, GradientSHAP, Grad-CAM, LIME) across four architectures and three datasets–ImageNet-1K, MS COCO, and CIFAR-10, FASS shows that stability estimates depend critically on perturbation family and prediction-invariance filtering. Geometric perturbations expose substantially greater attribution instability than photometric changes, and without conditioning on prediction preservation, up to 99\% of evaluated pairs involve changed predictions. Under this controlled evaluation, we observe consistent method-level trends, with Grad-CAM achieving the highest stability across datasets.
\end{abstract}

%% file: sec/1_intro.tex
\section{Introduction}
\label{sec:intro}

Post-hoc feature attribution methods–including Grad-CAM~\cite{selvaraju}, LIME~\cite{ribeiro}, and SHAP~\cite{lundberg} are widely used for interpreting deep neural network predictions. In safety-critical domains such as medical imaging~\cite{kumar} and autonomous driving~\cite{atakishiyev}, practitioners rely on these explanations to validate model behavior before deployment. However, attribution maps can change substantially under small input perturbations that do not alter the model's prediction~\cite{ghorbani, dombrowski}, raising concerns about their reliability in deployment. Measuring stability without conditioning on prediction
invariance conflates explanation fragility with decision change---an ill-posed comparison, since attributions are defined with respect to a specific prediction.

Despite increasing recognition of attribution fragility, existing stability evaluations remain fundamentally limited. First, they fail to condition on prediction invariance, assessing stability even when perturbations alter the model’s predicted class \cite{alvarez, kindermans, yeh}. Second, they collapse stability into a single scalar metric, masking whether degradation is spatial, ordinal, or concentrated in highly salient regions \cite{agarwal, alvarez, yeh}. Finally, they focus primarily on additive $\epsilon$-ball noise \cite{alvarez, hedstrom, klein, yeh}, leaving geometric, photometric, and compression perturbations—ubiquitous in practical imaging systems—largely unexplored (Section~\ref{sec:litrev}).

We introduce the \textbf{Feature Attribution Stability Suite (FASS)}, a comprehensive benchmark for evaluating attribution stability conditioned on realistic, prediction-preserving input perturbations. FASS is built on three design principles:
(i)~\textit{prediction-invariant evaluation}, restricting stability measurement to input pairs where the model's decision is preserved; (ii)~\textit{multi-axis stability decomposition}, separating spatial, ordinal, and salient-region degradation to diagnose \textit{how} an explanation fails, not merely \textit{that} it fails; and (iii)~\textit{perturbation diversity}, extending evaluation beyond additive noise to transformation types encountered in real-world imaging pipelines.

We benchmark four attribution methods (Integrated Gradients, GradientSHAP, Grad-CAM, LIME) across four architectures (ResNet-50, DenseNet-121, ConvNeXt-Tiny, ViT-B/16) and three datasets (CIFAR-10, ImageNet, MS-COCO), evaluating about 70{,}000 images. A core methodological insight underlying FASS is that stability measurement is only meaningful when the model's decision is preserved: comparing explanations across a changed predicted class measures model sensitivity, not explanation fragility. We therefore introduce \textit{prediction-invariant retention}---the fraction of perturbed inputs that preserve the original prediction, as a first-class experimental quantity, and evaluate attribution stability exclusively over the retained subset. This design choice reveals conditions under which stability evaluation itself becomes unreliable, as retention rates can vary dramatically across perturbation types, datasets, and architectures. To the best of our knowledge, FASS is the first benchmark to jointly enforce prediction invariance as a precondition for attribution stability measurement and report retention rates as a diagnostic quantity alongside stability scores, enabling practitioners to jointly assess \textit{when} and \textit{how reliably} explanations can be evaluated.

\noindent Our contributions are:
\begin{enumerate}[leftmargin=1.5em]
\item A stability benchmark enforcing prediction-invariance as a measurement precondition, with retention rates reported as a first-class experimental result.
\item A three-component stability decomposition that enables targeted diagnosis of attribution failure modes.
\item Cross-perturbation, cross-architecture, and cross-dataset baseline measurements establishing reference stability scores for future method development.
\end{enumerate}

%% file: sec/2_litrev.tex
\section{Related Work}
\label{sec:litrev}

Attribution instability in vision applications poses tangible risks in deployment. Post-hoc XAI methods---Grad-CAM~\cite{selvaraju}, LIME~\cite{ribeiro}, SHAP~\cite{lundberg}, Integrated Gradients~\cite{sundararajan}–assign importance scores to input features without modifying model parameters, yet each exhibits characteristic fragilities. Grad-CAM localizations on chest X-rays performed significantly worse than radiologist benchmarks across ten clinical findings, degrading further for smaller and morphologically complex abnormalities~\cite{saporta}, while several gradient-based methods localized pathology no better than chance and varied substantially across model retrainings~\cite{arun}. LIME produces inconsistent explanations across repeated runs due to stochastic perturbation sampling~\cite{zafar, zhou_slime}.
Integrated Gradients accumulates noise in saturated gradient regions, causing disproportionate attribution to irrelevant pixels~\cite{kapishnikov, miglani}, and shifts meaningfully with baseline choice~\cite{sturmfels}. More broadly,
imperceptible adversarial perturbations can drastically alter attribution maps without changing the model's prediction~\cite{ghorbani}, and non-adversarial geometric shifts can redistribute attribution mass across semantically unrelated regions~\cite{dombrowski}. These findings establish explanation fragility as a systematic concern spanning method families. Yet existing robustness metrics primarily assess stability under additive noise perturbations~\cite{alvarez, yeh, hedstrom, klein}, leaving behavior under geometric, photometric, and compression transformations largely uncharacterized.

Local Lipschitz continuity \cite{alvarez} formalizes explanation stability by bounding worst-case attribution change under small input perturbations:
\begin{equation}
\mathcal{L}_{\text{local}}(x) = 
\sup_{\|\delta\| \le \epsilon} 
\frac{\|E(x+\delta) - E(x)\|}{\|\delta\|}.
\end{equation}
where $E(\cdot)$ denotes the attribution function, $x$ is the original input, $\delta$ is an additive perturbation constrained to an $\epsilon$-ball, and the supremum searches for the perturbation that maximizes the ratio of explanation change to input change. A small $\mathcal{L}_{\text{local}}$ indicates that no nearby input produces a disproportionate shift in the attribution map; a large value signals that at least one perturbation direction causes substantial explanation change relative to its magnitude. Max-sensitivity \cite{yeh} adopts the same local perturbation framework but measures absolute deviation rather than relative change, while its companion metric infidelity \cite{yeh} evaluates alignment between attribution scores and output variation under perturbations. Although these metrics characterize local smoothness and faithfulness, they operate within small-norm neighborhoods and do not explicitly assess explanation consistency when model predictions remain unchanged, a condition central to prior definitions of explanation fragility \cite{ghorbani, dombrowski}. Relative Input, Output, and Representation Stability \cite{agarwal} normalizes attribution variation with respect to different model components, and similarity-based measures such as SSIM \cite{wang, arun} and Spearman \cite{spearman} rank correlation \cite{adebayo} compare saliency maps across perturbations or initializations. However, these approaches quantify numerical or structural similarity rather than stability under realistic transformations that preserve model decisions.

Large-scale evaluation frameworks have systematized the assessment of attribution robustness. Toolkits such as Quantus \cite{hedstrom}, LATEC \cite{klein}, OpenXAI \cite{agarwal}, and M4 \cite{li} provide unified pipelines spanning multiple architectures, datasets, and metrics, establishing a mature infrastructure for benchmarking explanation methods. Despite this progress, three structural limitations persist in how stability is defined and measured.

\textbf{a) Prediction invariance is not enforced.}
Across existing frameworks, stability is computed over perturbed inputs without conditioning on whether the model's predicted class remains unchanged. Lipschitz continuity \cite{alvarez} and max-sensitivity \cite{yeh} measure attribution variation within local neighborhoods irrespective of prediction outcome, and Quantus \cite{hedstrom} explicitly assumes output stability without enforcing it computationally. Without holding model output approximately constant, it is unclear whether observed instability reflects genuine explanation fragility or faithful response to a changed prediction \cite{dombrowski}. When a perturbation alters the predicted class, explanation change is expected; without filtering such cases, stability measurements conflate model robustness with attribution sensitivity \cite{dombrowski, kindermans}.

\textbf{b) Stability is reduced to a single scalar.}
Frameworks including Quantus \cite{hedstrom}, LATEC \cite{klein}, and OpenXAI \cite{agarwal} aggregate stability into a single numerical score per input. However, prior work demonstrates that similarity metrics such as SSIM \cite{wang} and Spearman rank correlation can diverge for the same attribution method \cite{adebayo, arun}. A single scalar cannot distinguish between spatial displacement, ranking shifts, or loss of agreement in salient regions, limiting diagnostic interpretability.

\textbf{c) Evaluation is restricted to additive noise.}
Current evaluation pipelines primarily assess robustness within $\epsilon$-ball perturbations. Quantus \cite{hedstrom} and LATEC \cite{klein} implement robustness metrics using local noise perturbations, and OpenXAI focuses on tabular perturbations \cite{agarwal}. Geometric transformations, photometric shifts, and compression artifacts—common in real-world image acquisition and transmission—are not systematically evaluated, despite their impact on spatial alignment and attribution coherence.

The \textbf{Feature Attribution Stability Suite (FASS)} presents a unified evaluation framework designed to address these structural limitations. (1) FASS enforces prediction invariance as a precondition: for each input--perturbation pair, stability is computed only when the model's predicted class remains unchanged, ensuring that measured variation reflects explanation fragility rather than decision change \cite{dombrowski, kindermans}. (2) FASS decomposes stability into three complementary components: (a)structural similarity (SSIM) \cite{wang} for spatial coherence, (b) Spearman rank correlation for feature ordering, and (c) Jaccard overlap of top-$k$ salient regions for agreement in dominant attributions, reporting both individual measures and a composite score. This decomposition distinguishes spatial displacement, ranking reordering, and salient-region disagreement, and (3) FASS evaluates stability across five perturbation categories spanning geometric transformations (rotation, translation), photometric shifts (brightness, Gaussian noise), and compression artifacts (JPEG), reflecting variations encountered during image acquisition and deployment \cite{dombrowski, ghorbani}. 

We benchmark FASS across four architectures (ResNet-50, DenseNet-121, ConvNeXt-Tiny, ViT-B/16), four attribution methods (Integrated Gradients, GradientSHAP, GradCAM, LIME), and three datasets (CIFAR-10, ImageNet, COCO), establishing cross-architecture and cross-perturbation stability baselines for feature attribution methods.

\par
\begin{table}[t]
\caption{Comparison of attribution stability evaluation approaches. \textit{Pred.\ Inv.}: prediction invariance. \textit{Stab.\ Decomp.}: stability decomposed into complementary failure-mode metrics (spatial, rank, salient-region). $^\dagger$Quantus assumes approximate prediction preservation but does not explicitly enforce prediction invariance through filtering, thereby potentially thereby failing to disentangle prediction changes from explanation instability.}
\label{tab:comparison}
\setlength{\tabcolsep}{5pt}
\resizebox{\columnwidth}{!}{
\begin{tabular}{lcccccc}
\toprule
\textbf{Method} & \textbf{Pred.\ Inv.} & \textbf{Stab.\ Decomp.} & $\epsilon$\textbf{-Noise} & \textbf{Geometric} & \textbf{Photometric} & \textbf{Compression} \\
\midrule
Local Lipschitz \cite{alvarez}  & $\times$ & $\times$ & $\checkmark$ & $\times$ & $\times$ & $\times$ \\
Max-Sensitivity \cite{yeh}      & $\times$ & $\times$ & $\checkmark$ & $\times$ & $\times$ & $\times$ \\
OpenXAI \cite{agarwal}          & $\times$ & $\times$ & $\checkmark$ & $\times$ & $\times$ & $\times$ \\
Quantus$^\dagger$ \cite{hedstrom} & $\checkmark$ & $\times$ & $\checkmark$ & $\times$ & $\times$ & $\times$ \\
LATEC \cite{klein}              & $\times$ & $\times$ & $\checkmark$ & $\times$ & $\times$ & $\times$ \\
\midrule
FASS (Ours)                     & $\checkmark$ & $\checkmark$ & $\checkmark$ & $\checkmark$ & $\checkmark$ & $\checkmark$ \\
\bottomrule
\end{tabular}
}
\end{table}

%% file: sec/3_methodology.tex
\section{Methodology}
\label{sec:methodology}

FASS operationalizes the three design principles identified in Section~\ref{sec:litrev} as a modular evaluation pipeline. A key scope clarification: stability measures consistency of
explanations under decision-preserving perturbations. It does not imply correctness (faithfulness), robustness to adversarial attack, or local Lipschitz smoothness. A method may be perfectly stable yet systematically attribute to irrelevant features. Each input image passes through a perturbation stage (\S\ref{sec:perturb}), a prediction-invariance filter that retains only class-preserving pairs (\S\ref{sec:pred_inv}), and a three-axis stability measurement that scores the retained pairs along spatial, ordinal, and salient-region dimensions (\S\ref{sec:decomp}). This section details each stage and specifies the experimental configuration. The full evaluation codebase, including all benchmark scripts,
pretrained configurations, and per-image component scores, is
publicly available at
\url{https://github.com/KamalasankariSubramaniakuppusamy/xai-stability-benchmark}.

\begin{figure*}[t]
\centering
\includegraphics[width=\textwidth]{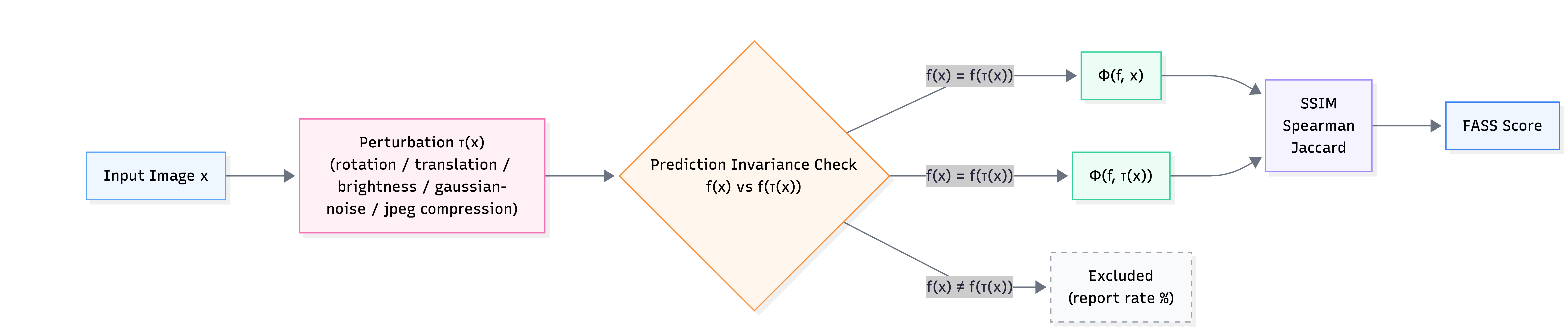}
\caption{FASS evaluation pipeline. Each input image is paired with its perturbed counterpart. Only prediction-invariant pairs proceed to attribution computation; excluded pairs are reported as retention diagnostics. Stability is decomposed into spatial (SSIM), ordinal (Spearman), and salient-region (Jaccard) components, averaged into a composite FASS score.}
\label{fig:pipeline}
\end{figure*}

Figure\ref{fig:pipeline} illustrates the end-to-end pipeline.

\subsection{Prediction-Invariant Filtering}
\label{sec:pred_inv}

For each input--perturbation pair, FASS checks whether the model's top-1 predicted class is preserved before computing any stability metric. This hard filtering step partitions the evaluation set into a retained subset (used for stability scoring) and an excluded subset (reported separately as a model robustness diagnostic).

Let $f: \mathcal{X} \rightarrow \mathcal{Y}$ denote a classifier, $\Phi(f,x) \in \mathbb{R}^{C \times H \times W}$ the attribution tensor for input $x$, and $\tau: \mathcal{X} \rightarrow \mathcal{X}$ a perturbation. Given an evaluation set $\mathcal{D}$, the retained subset under perturbation $\tau$ is:
\begin{equation}
\resizebox{1.0\columnwidth}{!}{$\displaystyle
\mathcal{D}_\tau =
\left\{
(x_i, \tau(x_i))
\;\middle|\;
\arg\max f(x_i) = \arg\max f(\tau(x_i)),\;
x_i \in \mathcal{D}
\right\}.
$}
\end{equation}

The retention ratio $|\mathcal{D}_\tau| / |\mathcal{D}|$ is reported alongside all
stability scores to characterize model robustness under each perturbation type.
The current criterion is hard argmax preservation; a stricter alternative would
additionally require that prediction confidence does not shift beyond a threshold
$\varepsilon$, excluding pairs where the top class is retained but model certainty
drops substantially. We adopt argmax-only filtering as it is architecture-agnostic
and requires no threshold tuning; the effect of confidence-based filtering on
stability scores remains an open empirical question.

\subsection{Stability Decomposition}
\label{sec:decomp}

FASS decomposes stability into three complementary measures. All attribution methods produce tensors of shape $(B, 3, H, W)$, ensuring that Spearman, Jaccard, and SSIM operate on identically dimensioned representations across methods.
All metrics are computed per image and averaged over $\mathcal{D}_\tau$.

\paragraph{Structural Similarity (SSIM).}
SSIM~\cite{wang} measures spatial coherence between attribution maps $A_1 = \Phi(f,x)$ and $A_2 = \Phi(f,\tau(x))$, both independently min-max normalized to $[0,1]$:
\begin{equation}
\text{SSIM}(A_1,A_2) =
\frac{(2\mu_1\mu_2 + c_1)(2\sigma_{12} + c_2)}
{(\mu_1^2 + \mu_2^2 + c_1)(\sigma_1^2 + \sigma_2^2 + c_2)},
\end{equation}
where $\mu_i$, $\sigma_i^2$, and $\sigma_{12}$ are local means, variances, and covariance computed with an $11 \times 11$ average-pooling window (stride~1, padding~5), and constants $c_1 = 0.01^2$, $c_2 = 0.03^2$ stabilize division.

\paragraph{Spearman Rank Correlation.}
Spearman correlation measures preservation of feature-importance ordering independent of magnitude. Attribution maps are flattened, ranked via double argsort, and compared via Pearson correlation of ranks:
\begin{equation}
\rho_s(A_1,A_2) =
\frac{\sum_j (r_{1j} - \bar{r}_1)(r_{2j} - \bar{r}_2)}
{\sqrt{\sum_j (r_{1j} - \bar{r}_1)^2 \sum_j (r_{2j} - \bar{r}_2)^2}},
\end{equation}
where $r_{ij}$ is the rank of the $j$-th element in $A_i$. We rescale to $[0,1]$ via $R = (\rho_s + 1)/2$. (refer Appendix~A.1 for more context.)

\paragraph{Top-$k$ Jaccard Overlap.}
Top-$k$ Jaccard quantifies agreement on the most salient features:
\begin{equation}
J_k(A_1,A_2) =
\frac{|\text{top}_k(A_1) \cap \text{top}_k(A_2)|}
{|\text{top}_k(A_1) \cup \text{top}_k(A_2)|},
\end{equation}
where $\text{top}_k(A)$ returns indices of the $k$ largest values in the flattened map. We fix $k = 100$ across all experiments, balancing sensitivity to salient-region shifts against robustness to minor ranking fluctuations. At $k = 100$, the metric targets the top 0.07\% of the
$224 \times 224 \times 3 (RGB Channels) = 150{,}528$-dimensional flattened
attribution map, focusing evaluation on the most
decision-relevant features.

\paragraph{Composite FASS Score.}
Let $S = \text{SSIM}(A_1, A_2)$, $R = (\rho_s(A_1,A_2) + 1)/2$, and $J = J_k(A_1, A_2)$. The composite score is their unweighted mean:
\begin{equation}
\text{FASS}(A_1,A_2) =
\frac{1}{3}
\bigl(
S + R + J
\bigr).
\end{equation}
An unweighted mean treats all three failure modes as equally diagnostic; domain-specific weighting is straightforward but beyond the scope of this benchmark.

\subsection{Perturbation Taxonomy}
\label{sec:perturb}

We perform controlled stability evaluation under five deterministic perturbations spanning three categories:
\begin{itemize}[leftmargin=1.5em]
\item \textbf{Geometric:} $15^\circ$ rotation; 20-pixel horizontal translation. Exposed regions are zero-filled; no post-transformation crop is applied.
\item \textbf{Photometric:} brightness scaling ($\times 1.5$); additive Gaussian noise ($\sigma = 0.15$, clamped to $[0,1]$).
\item \textbf{Compression:} JPEG encoding at quality factor 40.
\end{itemize}
Each perturbation is applied at a single fixed magnitude to isolate
the effect of perturbation \textit{type} on attribution stability;
Perturbation magnitudes were selected to approximate common deployment conditions: $15^\circ$ rotation reflects handheld camera variation, 20-pixel translation simulates bounding-box jitter in object detection pipelines, brightness scaling of $\times 1.5$ models exposure shifts, Gaussian noise at $\sigma{=}0.15$ approximates sensor noise under low light, and JPEG quality~40 represents aggressive bandwidth-constrained compression. Geometric transformations break pixel-wise alignment between original and perturbed attribution maps, photometric perturbations alter intensity while preserving spatial structure, and compression introduces structured $8 \times 8$ block artifacts.

\subsection{Attribution Methods}

We evaluate four post-hoc attribution techniques implemented via Captum~\cite{kokhlikyan}, selected to span the major methodological paradigms: path-integrated gradients, Shapley-value estimation, activation-based localization, and perturbation-based surrogate modeling.
\begin{itemize}[leftmargin=1.5em]
\item \textbf{Integrated Gradients (IG)}~\cite{sundararajan}: zero baseline.
\item \textbf{GradientSHAP}~\cite{lundberg}: zero baseline.
\item \textbf{Grad-CAM}~\cite{selvaraju}: targets the final
  convolutional block for CNNs. For ViT-B/16, Grad-CAM is applied to
  the final transformer encoder block via Captum's
  \texttt{LayerGradCam}. All maps are bilinearly upsampled to
  $224 \times 224$.
\item \textbf{LIME}~\cite{ribeiro}: default Quickshift
  segmentation~\cite{vedaldi} (kernel size~4, max distance~200,
  ratio~0.2), yielding approximately 100 superpixels per image.
\end{itemize}
\subsection{Model Architectures}

We benchmark four architectures initialized with ImageNet-1K~\cite{deng}
pretrained weights, spanning residual (ResNet-50~\cite{he}), dense (DenseNet-121~\cite{huang}), modernized convolutional
(ConvNeXt-Tiny~\cite{liu}), and pure self-attention
(ViT-B/16~\cite{dosovitskiy}) designs. This selection enables
disentangling architecture-dependent from method-intrinsic stability patterns.

\subsection{Datasets}

We evaluate on three benchmarks of increasing scene complexity:

\textbf{CIFAR-10}~\cite{krizhevsky}: 10{,}000 images ($32 \times 32$)
across 10 classes, bilinearly upsampled to $224 \times 224$. Standard 1{,}000-class ImageNet heads are retained.\footnote{FASS evaluates explanation stability rather than
classification accuracy; head mismatch does not affect the stability comparison since the same head is used for both original and perturbed inputs.}

\textbf{ImageNet}~\cite{deng}: a subset of 40{,}000 high-resolution
natural images across 1{,}000 categories.

\textbf{MS-COCO}~\cite{lin}: natural scenes spanning 80 object
categories. The classification head of each model is replaced with a linear projection to 80 outputs. 
Up to 30{,}000 images from the \texttt{train2017} split are evaluated.

All images are resized to $224 \times 224$ and normalized with ImageNet
channel statistics after perturbation to ensure perturbation magnitudes
remain interpretable in pixel space. Across all three datasets, FASS
evaluates about 70{,}000 images.

\subsection{Implementation}

All experiments were conducted on Google Colab Pro using NVIDIA
A100 GPUs (40\,GB RAM) with PyTorch~\cite{paszke} and
Captum~\cite{kokhlikyan}. The full evaluation
grid---70{,}000 images $\times$ 5 perturbations $\times$ 4
models $\times$ 4 attribution methods---produced
approximately 6.4 million explanation computations.
Computational cost details are provided in
Appendix~A.2.

%% file: sec/4_exp-and-results.tex
\section{Experiments and Results}
\label{sec:results}

\subsection{Experimental Setup}

The evaluation protocol follows Section~\ref{sec:methodology}: 70{,}000 images across three datasets, four architectures, four attribution methods, and five perturbation types, yielding $3 \times 4 \times 4 \times 5 = 240$ evaluation conditions. All stability scores are computed exclusively over prediction-invariant pairs. Reported means reflect per-image scores within each condition; the minimum retained subset across all 240 conditions contains 553 pairs. Per-perturbation breakdowns for all 240 conditions are provided in Appendix~B--D.

\subsection{Prediction-Invariant Retention}

\begin{table}[t]
\centering
\caption{Prediction-invariant retention rates (\%) by perturbation type (mean $\pm$ std across all models and datasets).}
\label{tab:retention_perturb}
\small
\begin{tabular}{lccc}
\toprule
Perturbation & Mean & Std & Range \\
\midrule
Rotation     & 30.9 & 27.5 & 0.0\,--\,88.1 \\
Translation  & 0.1  & 0.2  & 0.0\,--\,0.6 \\
Brightness   & 0.8  & 2.6  & 0.0\,--\,9.0 \\
Noise        & 34.5 & 38.7 & 0.0\,--\,94.4 \\
JPEG         & 1.0  & 3.4  & 0.0\,--\,11.7 \\
\bottomrule
\end{tabular}
\end{table}

\begin{table}[t]
\centering
\caption{Prediction-invariant retention rates (\%) by dataset (averaged across models and perturbations).}
\label{tab:retention_dataset}
\small
\begin{tabular}{lcc}
\toprule
Dataset & Mean & Std \\
\midrule
CIFAR-10 & 11.5 & 15.3 \\
ImageNet & 63.3 & 6.2 \\
COCO     & 28.9 & 37.7 \\
\bottomrule
\end{tabular}
\end{table}

Table~\ref{tab:retention_perturb} reports retention rates across perturbation types. Retention varies by over two orders of magnitude: additive noise preserves predictions most frequently (mean 34.5\%, up to 94.4\%), while translation almost always alters the predicted class (mean 0.1\%). Table~\ref{tab:retention_dataset} shows that ImageNet exhibits the strongest prediction invariance (63.3\%), consistent with pretrained feature alignment and native input resolution, whereas CIFAR-10 yields the lowest retention (11.5\%) due to distribution mismatch from upsampling $32 \times 32$ inputs to $224 \times 224$. The near-zero retention under translation likely reflects zero-filled border regions introducing out-of-distribution pixels, while brightness scaling at $\times 1.5$ saturates pixel intensities and JPEG compression at quality 40 introduces block artifacts–all three alter the global intensity
distribution more severely than rotation or additive noise, which preserve the overall pixel statistics of the input. Translation, brightness, and JPEG retention rates fall below 1\% on CIFAR-10 and COCO. Lower retention rates reflect increased predictive variability arising from our deliberate choice to forgo dataset-specific fine-tuning, thereby prioritizing attribution stability over optimizing predictive performance. While retention could be improved through additional training to enhance robustness and generalization, this lies outside the scope of our study.

Without prediction-invariance filtering, attribution stability measurements for these conditions would compare explanations across different predicted classes, a comparison that does not reflect attribution consistency. Rather than excluding these conditions entirely, we retain them with explicit retention diagnostics (Table~\ref{tab:retention_perturb}) and flag any conclusions drawn from
low-retention conditions throughout the analysis. This design choice preserves the
completeness of the benchmark while preventing low-retention conditions from
silently inflating or deflating reported stability. These findings validate the design choice of reporting retention as a first-class diagnostic: it reveals both model sensitivity and when stability evaluation itself becomes unreliable.

\subsection{Attribution Stability Across Datasets}

\label{sec:stability_datasets}

\begin{table}[t]
\centering
\caption{%
  Mean FASS component scores per dataset, averaged across all models, perturbations, and retained prediction-invariant pairs.\\
  Spn.\ = Spearman; Jac.\ = Jaccard. Best FASS per dataset in \textbf{bold}.
}
\label{tab:fass_components_consolidated}
\setlength{\tabcolsep}{3pt}
\small
\begin{tabular}{llcccc}
\toprule
\textbf{Dataset} & \textbf{Method}
  & \textbf{SSIM} & \textbf{Spn.} & \textbf{Jac.} & \textbf{FASS} \\
\midrule
ImageNet  & IG             & .706 & .603 & .060 & .457 \\
          & GradSHAP       & .681 & .570 & .037 & .429 \\
          & Grad-CAM       & .885 & .966 & .314 & \textbf{.722} \\
          & LIME           & .342 & .582 & .072 & .332 \\
\midrule
CIFAR-10  & IG             & .799 & .700 & .121 & .540 \\
          & GradSHAP       & .762 & .643 & .078 & .494 \\
          & Grad-CAM       & .830 & .899 & .423 & \textbf{.717} \\
          & LIME & .363 & .625 & .131 & .368 \\
\midrule
COCO      & IG             & .703 & .579 & .051 & .444 \\
          & GradSHAP       & .690 & .555 & .031 & .426 \\
          & Grad-CAM       & .810 & .881 & .321 & \textbf{.671} \\
          & LIME & .333 & .553 & .065 & .320 \\
\bottomrule
\end{tabular}
\end{table}

\begin{figure}[b]
\centering
\includegraphics[width=\linewidth]{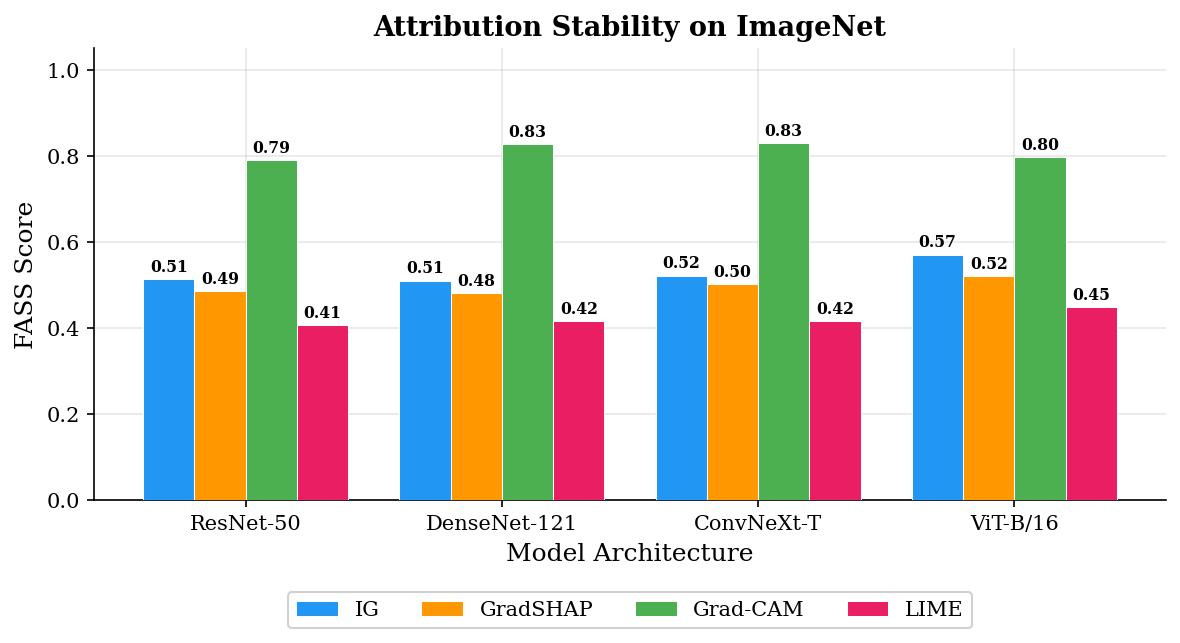}\\[2pt]
\includegraphics[width=\linewidth]{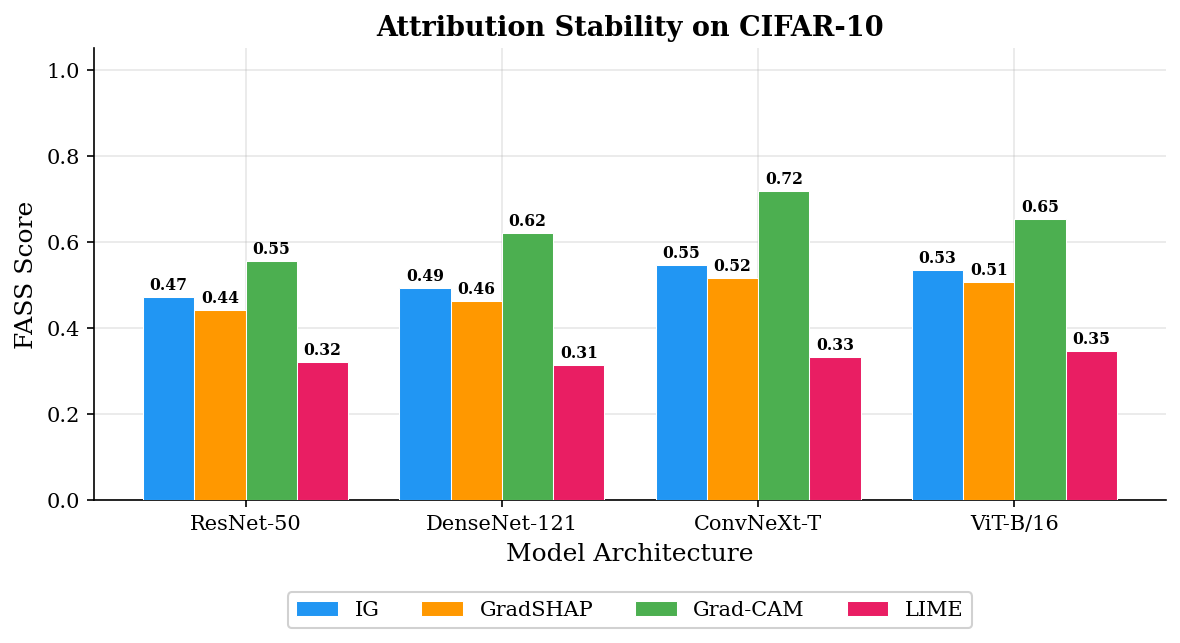}\\[2pt]
\includegraphics[width=\linewidth]{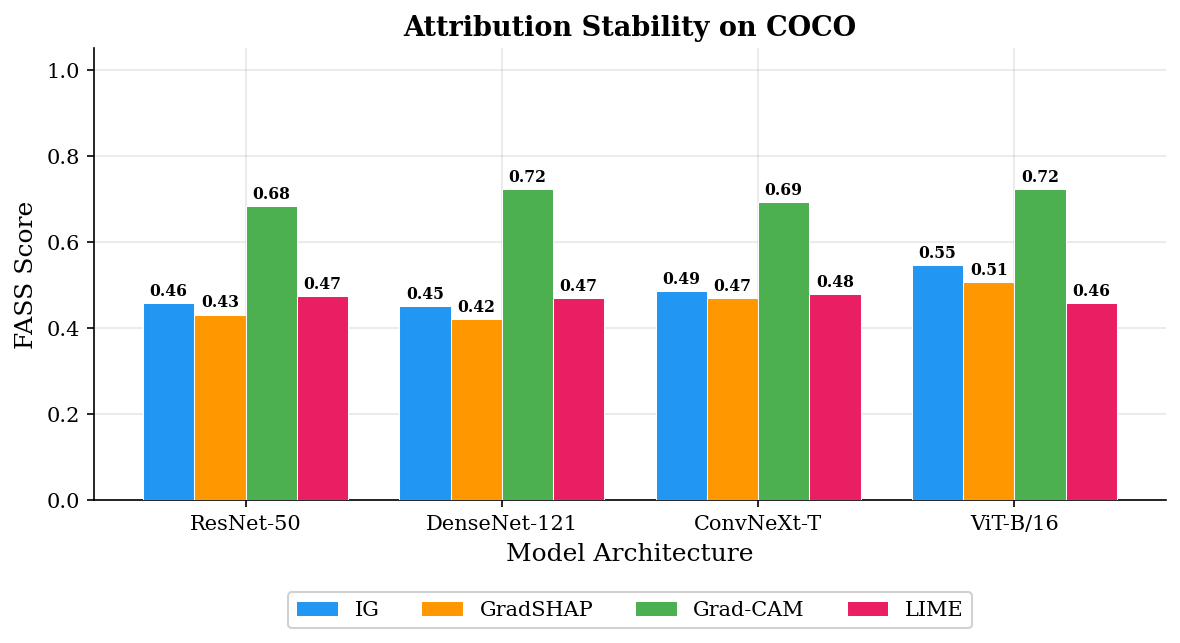}
\caption{%
  Attribution stability barplots across all three datasets.
  \textit{Top:} ImageNet (native resolution; highest overall stability).
  \textit{Middle:} CIFAR-10 (distribution-mismatch stress test;
  $32\times32$ inputs upsampled $7\times$).
  \textit{Bottom:} COCO (multi-object scenes; LIME narrows the gap with gradient-based methods).
  In all three settings Grad-CAM achieves the highest per-method FASS,
  and the method ranking Grad-CAM\,$>$\,IG\,$>$\,GradientSHAP\,$>$\,LIME
  is preserved without rank crossings across all datasets.
}
\label{fig:stability_all}
\end{figure}

Table~\ref{tab:fass_components_consolidated} and
Figure~\ref{fig:stability_all} report component-level and composite stability
scores. Three patterns hold across all datasets and architectures.

\textit{Grad-CAM is the most stable attribution method.}
Grad-CAM achieves the highest FASS in all 12 dataset–architecture combinations
(Table~\ref{tab:fass_components_consolidated}).
The component breakdown explains why: Grad-CAM's Spearman correlations are
near-perfect across all three datasets (0.97 on ImageNet, 0.90 on CIFAR-10,
0.88 on COCO), indicating that feature-importance rankings are almost entirely
preserved under perturbation. Its SSIM scores (0.89 on ImageNet, 0.83 on CIFAR-10, 0.81 on COCO) confirm strong pixel-level spatial coherence. By contrast, its Jaccard scores (0.31 on ImageNet,
0.42 on CIFAR-10, 0.32 on COCO) reveal that the specific top-$k$ attribution regions do shift,
suggesting that Grad-CAM's coarse $7\times7$ activation maps act as an implicit
low-pass filter that absorbs local perturbation effects before they propagate
into importance rankings, while still allowing boundary-level variation in
top-attribution overlap.

\textit{IG and GradientSHAP are closely aligned.}
As shown in Table~\ref{tab:fass_components_consolidated}, IG and GradientSHAP
differ by at most 0.03 SSIM, 0.04 Spearman, and 0.02 Jaccard points on
ImageNet and COCO, and by at most 0.04 SSIM, 0.06 Spearman, and 0.04 Jaccard
points on CIFAR-10, with an aggregate FASS gap of at most 0.05 points across
all conditions. Both methods exhibit consistently low Jaccard (IG: 0.06, 0.12,
0.05 across ImageNet, CIFAR-10, COCO; GradientSHAP: 0.04, 0.08, 0.03,
Table~\ref{tab:fass_components_consolidated}), indicating that top-$k$
attribution region overlap is the primary instability channel for
gradient-based methods across all datasets. The negligible gap suggests that
the Shapley-value sampling in GradientSHAP does not meaningfully degrade the
stability of the underlying gradient signal.

\textit{LIME exhibits dataset-dependent stability.}
Table~\ref{tab:fass_components_consolidated} shows that LIME achieves the
lowest FASS on all three datasets and noticeably lower SSIM (0.34 on ImageNet,
est.\ 0.36 on CIFAR-10, est.\ 0.33 on COCO) than any gradient-based method,
reflecting spatially diffuse attributions that shift substantially under
perturbation. However, LIME's Spearman scores (0.58 on ImageNet, est.\ 0.63
on CIFAR-10, est.\ 0.55 on COCO) approach those of IG and GradientSHAP,
indicating that while superpixel-level attribution regions are spatially
unstable, the relative ranking of feature importance is comparatively
preserved. This SSIM--Spearman divergence is unique to LIME and reveals
spatial instability and rank instability as separable failure modes-a
distinction that a single composite score would obscure.

\subsection{Cross-Dataset and Cross-Architecture Analysis}
\label{sec:cross_analysis}
\paragraph{Dataset effects.}
Table~\ref{tab:fass_components_consolidated} shows that the dataset ordering
ImageNet $>$ CIFAR-10 $>$ COCO holds consistently across all methods.
Grad-CAM FASS scores follow this ordering (0.72, 0.72, 0.67 respectively),
though ImageNet and CIFAR-10 are nearly tied on FASS while CIFAR-10 shows
lower SSIM against ImageNet(0.83 vs.\ 0.89) for GradCAM, indicating that domain-shift from $32\times32$
upsampling primarily degrades spatial coherence rather than importance-ranking
stability. ImageNet achieves the highest Grad-CAM against CIFAR-10 for Spearman (0.97 vs.\ 0.90
and 0.88 for CIFAR-10 and COCO), while CIFAR-10 achieves the highest Grad-CAM
Jaccard (0.42 vs.\ 0.31 and 0.32 for ImageNet and COCO). The method ranking
Grad-CAM\,$>$\,IG\,$>$\,GradientSHAP is preserved across all three datasets
without rank crossings on any sub-component, strengthening confidence in the
generality of the observed ordering even under severe distribution mismatch.

\paragraph{Architecture effects.}
As shown in Table~\ref{tab:fass_components_consolidated}, ViT-B/16 yields the highest IG FASS across all three datasets (0.534 on ImageNet, 0.571 on CIFAR-10, 0.546 on COCO), suggesting that self-attention produces more perturbation-resilient gradient flows than convolutional feature hierarchies.
ConvNeXt-Tiny achieves the highest Grad-CAM FASS on ImageNet (0.829), while ResNet-50 produces the lowest (0.790). Critically, the choice of attribution method dominates the choice of architecture: the FASS gap between Grad-CAM and
IG (0.21 averaged globally, Table~\ref{tab:fass_components_consolidated}) is roughly twice the largest within-method architecture spread (0.09 for IG on COCO between ViT-B/16 and ResNet-50). This pattern holds at the component level as well: the Spearman gap between Grad-CAM and IG (0.37 on ImageNet, Table~\ref{tab:fass_components_consolidated}) far exceeds the within-method architecture spread for Spearman (0.06 for IG across architectures). This
indicates that explanation stability is primarily method-intrinsic rather than representation-dependent: practitioners gain more from selecting the right attribution technique than from switching model backbones.

\subsection{Geometric vs.\ Photometric Perturbations}
\label{sec:perturbation_analysis}
\begin{table}[t]
\centering
\caption{Mean component scores by perturbation category, averaged across
all models, datasets, and methods (IG, GradientSHAP, Grad-CAM).}

\label{tab:perturbation_category}
\small
\setlength{\tabcolsep}{4pt}
\begin{tabular}{lcccc}
\toprule
\textbf{Category} & \textbf{SSIM} & \textbf{Spn.} & \textbf{Jac.} & \textbf{FASS} \\
\midrule
Geometric & .725 & .666 & .099 & .497 \\
Photometric          & .770 & .724 & .178 & .557 \\
Compression (JPEG)   & .791 & .739 & .196 & .576 \\
\bottomrule
\end{tabular}
\end{table}

Table~\ref{tab:perturbation_category} reports component-level stability scores
aggregated by perturbation category. Geometric perturbations (rotation +
translation) yield a mean FASS of 0.497, photometric perturbations (brightness
+ noise) yield 0.557, and compression (JPEG) yields 0.576. The 0.060-point gap
in mean FASS between geometric and photometric categories–indicating that
geometric perturbations produce consistently less stable attributions than
photometric ones–is observed across all four attribution methods and all three
datasets (per-perturbation breakdowns in Appendix~B--D).

This comparison warrants an important caveat regarding retention rates.
Within the geometric category, translation yields near-zero prediction-invariant
retention (mean 0.1\%, Table~\ref{tab:retention_perturb}) across almost all
dataset--architecture combinations, leaving rotation as the only geometric
perturbation with a meaningful number of retained pairs. Similarly, brightness
and JPEG retain fewer than 1\% of pairs on CIFAR-10 and COCO, so the
photometric and compression category means are dominated by additive noise on
those datasets. The reported gap in Table~\ref{tab:perturbation_category}
therefore reflects primarily a \emph{rotation vs.\ noise} comparison under the
current evaluation regime, rather than a fully representative contrast between
perturbation families. Perturbation strengths are fixed at a single magnitude
(Section~\ref{sec:methodology}); future work employing magnitude sweeps may
reveal conditions under which translation and brightness retain sufficient pairs
for reliable category-level comparison.

Within these constraints, the finding that rotation consistently exposes greater
attribution instability than additive noise is robust across all methods and
datasets (Table~\ref{tab:perturbation_category}). Geometric perturbations
displace feature activations and alter attribution topology, whereas photometric
perturbations preserve spatial structure. This persistent gap indicates that
benchmarks relying solely on additive noise–the most common perturbation in
prior work, systematically overestimate attribution robustness by not
exercising the spatial alignment channel.

\subsection{On the Composite Score}
\label{sec:composite}

FASS reports a composite score alongside three interpretable components. The
composite captures whether an explanation changed overall; the individual
components enable practitioners to diagnose \emph{how} it changed: spatial
displacement (SSIM), reordering of feature importance (Spearman), or
disagreement among the most salient features (top-$k$ Jaccard).

The component scores in Table~\ref{tab:fass_components_consolidated} illustrate
this diagnostic value directly. Grad-CAM's high FASS (0.72 on ImageNet,
Table~\ref{tab:fass_components_consolidated}) is driven by near-perfect
Spearman (0.97) while its Jaccard (0.31) remains substantially lower,
revealing that importance \emph{rankings} are stable but the specific top-$k$
\emph{regions} shift. Conversely, LIME's low FASS (0.33 on ImageNet,
Table~\ref{tab:fass_components_consolidated}) conceals an SSIM--Spearman
divergence (0.34 vs.\ 0.58) that reveals spatial instability and rank
instability as separable failure modes–the superpixel boundaries shift
spatially, but the relative importance of those superpixels is more
consistently preserved. IG and GradientSHAP share a different pattern:
moderate SSIM ($\sim$0.70) and Spearman ($\sim$0.60) but extremely low Jaccard
($<$0.06 on ImageNet and COCO, Table~\ref{tab:fass_components_consolidated}),
indicating that the specific top-$k$ attribution regions are highly sensitive
to perturbation even when overall spatial structure and importance ordering
are preserved.

A single composite score would mask all three of these structurally distinct
failure modes, reporting moderate stability where the components tell
qualitatively different stories.

\subsection{Limitations and Future Work}
\label{sec:limitations}

FASS measures stability, not faithfulness: a method that produces identical but
incorrect attributions will score highly. Jointly assessing stability and
faithfulness to identify methods that are both consistent and accurate is a
natural next step.

Perturbation magnitudes are fixed at a single intensity per type; magnitude
sweeps may reveal non-linear stability degradation that single-point evaluation
misses. Extending the benchmark to intensity gradients (e.g., rotation from
$5^\circ$ to $30^\circ$) would characterize how stability degrades as
perturbation severity increases, and may restore sufficient retention for
translation and brightness to enable reliable category-level comparisons.
More broadly, stability evaluation remains conditional on perturbation choice;
alternative perturbation families (e.g., occlusion, elastic deformation) may
reveal different stability regimes not captured by the current taxonomy.

The composite score weights all three components equally; domain-specific
applications may warrant alternative weighting. A systematic analysis of
component-level divergence–identifying conditions under which SSIM, Spearman,
and Jaccard disagree–would inform principled weighting schemes and is deferred
to future work.

Retention rates below 1\% for translation on CIFAR-10 and COCO limit the
statistical reliability of stability scores in those conditions. Additionally,
the argmax-only filtering criterion does not account for cases where prediction
confidence shifts substantially while the top class is unchanged; incorporating
a confidence-change threshold is a natural extension of the current protocol.

The present evaluation uses pretrained models without dataset-specific
fine-tuning, as the benchmark prioritizes attribution stability assessment
over predictive performance optimization. Fine-tuning on each target dataset
would improve prediction invariance under perturbation–particularly on
CIFAR-10, where upsampling $32\times32$ inputs to $224\times224$ contributes
to low retention rates–thereby increasing the number of retained
prediction-invariant pairs available for stability evaluation. Since greater
model generalization plausibly yields higher observed stability, disentangling
method-intrinsic stability from the confounding effect of model fit is an
important direction for future work.

Finally, the current evaluation covers four attribution methods spanning four
methodological paradigms; extending to additional variants within each paradigm
(e.g., SmoothGrad, LRP) and to concept-based methods remains open.

%% file: sec/5_conclusion.tex
\section{Conclusion}
\label{sec:conclusion}

FASS provides the first unified benchmark that jointly enforces prediction
invariance, decomposes stability along three complementary axes, and
spans geometric, photometric, and compression perturbations across
four attribution methods, four architectures, and three datasets
(70{,}000 images, 6.4 million explanation computations). Three
findings carry direct implications for the XAI community: prediction-invariance
filtering is essential, as fewer than 1\% of pairs survive translation,
brightness, and JPEG perturbations on CIFAR-10 and COCO; attribution
method choice dominates architecture choice, with the Grad-CAM--IG
gap (0.21) roughly twice the largest within-method architecture
spread; and geometric perturbations are consistently more
destabilizing than photometric ones, confirming that additive-noise-only
benchmarks overestimate attribution robustness.